\pdfoutput=1

\documentclass[11pt]{article}

\usepackage[]{acl}

\usepackage{times}
\usepackage{latexsym}
\usepackage{booktabs}
\usepackage{multirow}
\usepackage{amsmath}
\usepackage{amssymb}
\usepackage{graphicx}
\usepackage{xcolor}
\usepackage{url}
\usepackage{lipsum}
\usepackage{arydshln}
\usepackage{algorithm}
\usepackage{algpseudocode}
\usepackage{enumitem}

\usepackage[T1]{fontenc}

\usepackage[utf8]{inputenc}

\usepackage{microtype}

\title{LedgerAgent: Structured State for Policy-Adherent Tool-Calling Agents}

\author{Md Nayem Uddin$^{\spadesuit}$\hspace{9pt}
Amir Saeidi$^{\spadesuit}$\hspace{9pt}
Eduardo Blanco$^{\diamondsuit}$ \hspace{9pt}
Chitta Baral$^{\spadesuit}$ \\
$^\spadesuit$Arizona State University \; $^\diamondsuit$ University of Arizona \\
\texttt{muddin11@asu.edu} \\
}

\begin{document}
\maketitle
\begin{abstract}

Policy-adherent tool-calling agents in customer-service domains must maintain
task states across turns while calling tools and obeying domain policies. Task
states consist of relevant facts, identifiers, constraints, and conditions
observed through user interaction and tool calls.
In standard agents, task states are not represented separately.
Observations, tool returns, and policy instructions are placed in the prompt,
leaving agents to reconstruct the relevant states from the prompt each time they
decide what to do next.
This design makes state management implicit, creating two common failure modes.
An agent may retrieve the right facts but later ground its decision in stale,
missing, or incorrect information; and a syntactically valid tool call may still
violate a domain policy that depends on the current task state.
We introduce \textsc{LedgerAgent}, an inference-time method for tool-calling
agents that maintains observed task states in a separate ledger and renders the
states into the prompt.
The ledger is also used to check state-dependent policy constraints before
environment-changing tool calls are executed, blocking policy violations.
Across four customer-service domains and a mixed panel
of open- and closed-weight models, \textsc{LedgerAgent} improves average
pass\textasciicircum{}k over a standard prompt-based tool-calling approach, 
with the largest
gains under stricter multi-trial consistency metrics.

\end{abstract}

\section{Introduction}
\label{sec:introduction}

Language agents are increasingly evaluated in settings that require sustained
interaction rather than isolated tool calls~\citep{li2023apibank,liu2024agentbench,zhou2023webarena,lu2024toolsandbox,trivedi2024appworld}.
They must converse with users, retrieve records from external systems, 
and follow domain-specific rules across multiple turns.
Customer-service domains~\citep{yao2024tau,barres2025tau2} 
make this requirement concrete:
an agent may inspect a reservation, check an order, change a service plan, issue
a refund, or update an account. Success therefore depends on more than selecting
the right tool. The agent must maintain the relevant interaction state and act
only when the domain policy permits the action.

Most tool-calling agents expose information to the model through
prompt text~\citep{yao2023react,karpas2022mrkl,lu2023chameleon}.
Tool outputs are appended to the prompt. Prior actions remain
interleaved with user messages and model generations. The policy document is
supplied as natural-language instructions. At each turn, the model must identify
which prior facts matter, decide whether more information is needed, choose the
next response or tool call, and judge whether the intended action is allowed.
This design is simple and model-agnostic. However, it leaves task state implicit
in an ever-growing context, so reliable behavior depends on finding and using
the relevant evidence when the agent acts in the environment.

\begin{figure*}[t]
    \centering
    \includegraphics[width=\textwidth]{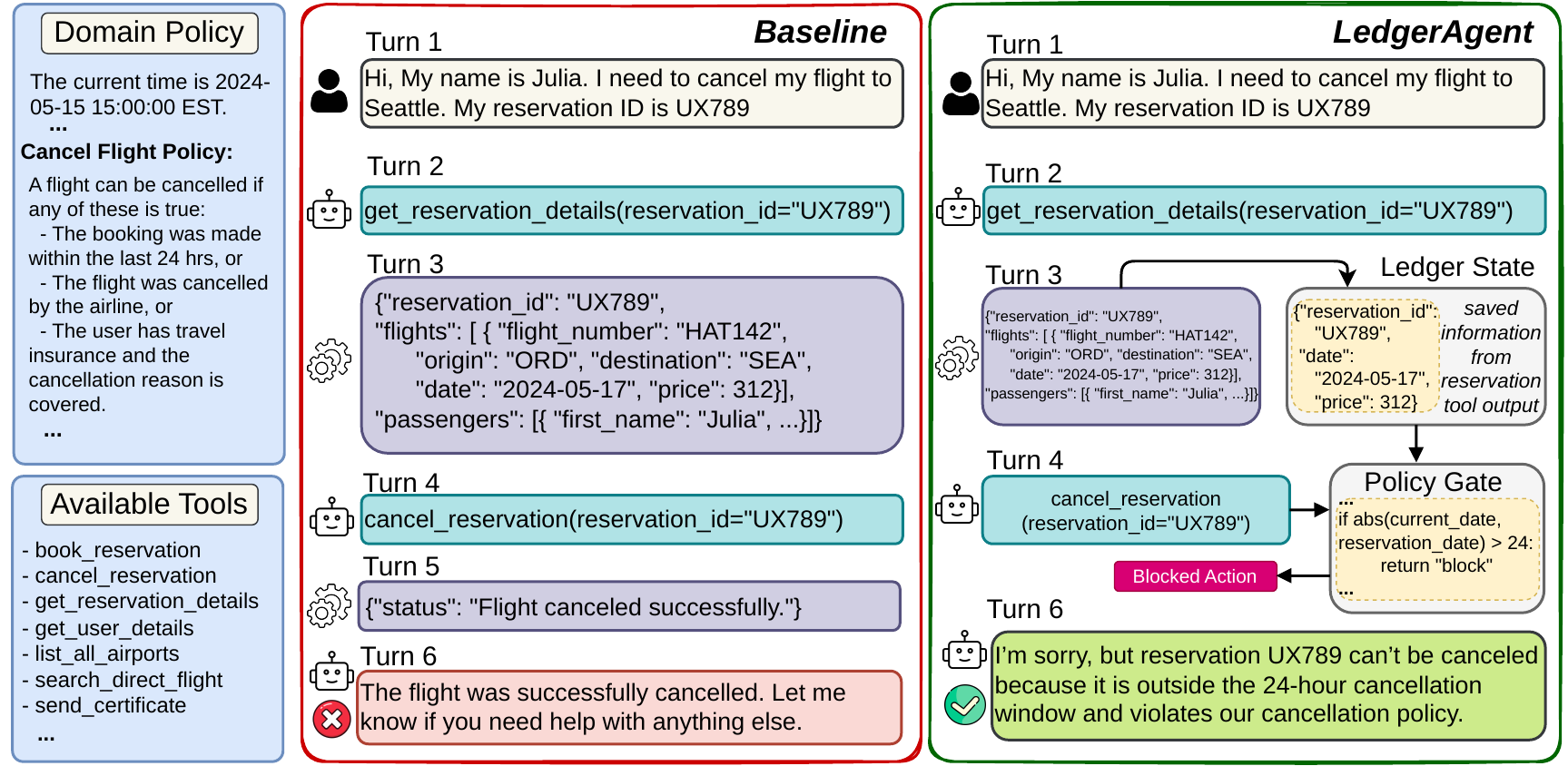}
    \caption{A standard agent retrieves a reservation record but later issues a policy-violating cancellation because the relevant state remains implicit in the prompt. LedgerAgent stores successful tool returns in a schema-anchored ledger and checks proposed environment-changing calls against this observed state before execution. In this example, the stored reservation state allows the gate to block the cancellation and return policy-grounded feedback.}
    \label{fig:teaser}
\end{figure*}


A second failure appears at the policy boundary. Domain policies specify when
actions are allowed. For example, they may define which orders are returnable or
which payment method must receive a refund. These rules are usually supplied
before the agent has retrieved the records that determine which rules 
apply~\citep{ruan2024toolemu,yao2024tau}.
When the agent later proposes an action, there is often no separate check against
the current state and governing policy. A tool call can therefore be syntactically valid while still violating the domain policy.

We introduce \textsc{LedgerAgent}, an inference-time method that adds an
explicit state representation in the agent loop. It has two deterministic
components. 
First, it maintains a schema-anchored ledger which projects
successful tool returns into a compact typed dictionary keyed by canonical paths. This requires no additional LLM calls to build.
The ledger is re-injected at each
turn, so the agent can consult current task state by lookup rather than searching
raw transcript history. Second, LedgerAgent applies a policy gate before
environment-changing tool calls are executed. The gate evaluates proposed calls
against domain rules expressed as predicates over ledger fields. If a proposed
call violates policy, the gate blocks it and returns feedback identifying the
violated rule and conflicting state. The agent can then block or revise its plan
before the environment is changed. Figure~\ref{fig:teaser} illustrates how explicit ledger state and a pre-action
policy gate change the failure boundary from post-hoc correction to prevention.

\textsc{LedgerAgent} addresses different failure modes from methods that primarily
improve the model. Fine-tuning and synthetic data generation teach models to
perform tool use more reliably~\citep{schick2023toolformer,li2023apibank,patil2023gorilla,qin2024toolllm}.
Reinforcement learning rewards successful trajectories~\citep{zhou2024archer,jin2025searchr1}.
Inference-time scaffolds add planning, reflection, or workflow constraints
around generation~\citep{yao2023react,yao2023tree,madaan2023selfrefine,shinn2023reflexion}.
These approaches can improve performance, but they largely preserve the same
prompt-only way of representing state. Retrieved records and policies remain
embedded in a growing transcript, and task state remains implicit at the moment
of action. LedgerAgent instead changes how state is represented. The model
weights are unchanged. The agent is given an explicit typed state object, and
policy is enforced when proposed actions are about 
to affect the environment.

We evaluate LedgerAgent across four customer-service domains from
$\tau^{2}$-bench and $\tau$-Trait~\citep{barres2025tau2,he2025impatient}, 
using a mixed panel of open- and closed-weight
models. LedgerAgent improves pass\textasciicircum{}k on majority of the domain--model pair we evaluated. The gains are
largest at higher values of $k$, where the standard prompt-based approach is
least consistent across independent trials. Ablations show that the improvement
comes from the typed ledger and policy-gated action. In summary, this paper makes three contributions:



\begin{itemize}
    \item We identify state grounding as a key failure mode in
    policy-adherent tool-calling agents: agents may retrieve the right records
    but later act on stale, missing, or incorrectly reconstructed state.

    \item We introduce \textsc{LedgerAgent}, an inference-time method that
    maintains observed state in a schema-anchored typed ledger, renders it for
    generation, and uses it to check environment-changing calls before
    execution.

    \item We show that \textsc{LedgerAgent} improves consistency-oriented
     pass\textasciicircum{}k across customer-service domains and backbone models,
    with the largest gains on tasks requiring environment-changing actions.
\end{itemize}
\section{Related Work}
\label{sec:related-work}

\paragraph{Interactive tool-using agents}
Recent work on language agents has moved beyond isolated API calls toward
interactive tasks in which models must use tools over several turns. Early
tool-use benchmarks and datasets study whether models can plan, select APIs, and
produce valid calls in tool-augmented settings \citep{li2023apibank}. More recent customer-service benchmarks make the setting
more realistic by combining dialogue, structured records, domain APIs, and
operational policies \citep{yao2024tau, barres2025tau2}. These
benchmarks reveal that agent failures are often not simple tool-selection
mistakes. A model can retrieve the right information and still make an incorrect
decision later because the relevant record is buried in the interaction history.

\paragraph{Inference-time scaffolding for tool use}
Many methods improve tool-using agents by changing the procedure around the
model rather than the model itself. Planning and reasoning frameworks encourage
the model to decompose tasks before acting, while reflection methods use feedback
from previous attempts to improve later behavior \citep{yao2023react,
shinn2023reflexion}. Recent input-reformulation and multi-agent approaches also
provide additional context or specialized assistance before the next tool call.
IRMA reformulates the agent input with relevant domain rules and tool
suggestions, while FAMA dynamically selects specialized helper agents based on
observed failure modes \citep{mishra2025irma, saeidi2026fama}. However, these
approaches still largely rely on the language model to recover the current task
state from the transcript, which can be unreliable when later actions depend on
exact records, identifiers, statuses, or valid tool arguments.

\paragraph{Policy adherence in tool-using agents}
A separate challenge is ensuring that tool calls satisfy domain constraints.
Prior benchmarks emphasize the importance of policy following in realistic
tool-agent-user interaction, but most agent implementations still place rules in
the prompt or rely on the model to reason about whether an action is allowed
\citep{yao2024tau, barres2025tau2}. This can fail when the
applicability of a rule depends on records obtained during the conversation.
LedgerAgent targets this gap at the interface between the model and the
environment. It gives the agent an explicit structured state object derived from
tool observations and uses that state to check proposed write actions before
execution.
Compared with methods that primarily rely on training, prompting, or multi-agent
orchestration, LedgerAgent frames state tracking and policy adherence as
system-level mechanisms that complement the model's reasoning.
\section{Method}
\label{sec:method}

\textsc{LedgerAgent} is an inference-time method for a standard
policy-adherent tool-calling agent. It adds two deterministic components: a
ledger that stores observed state from successful tool returns, and a
policy gate that checks environment-changing calls before execution.
Environment-changing calls modify external state, such as issuing refunds,
updating orders, changing reservations, or changing accounts; read-only calls are
not gated.

On each turn, new read-tool returns are absorbed into the ledger, the ledger is
rendered into the prompt, and the model generates a response or tool call. Before
any environment-changing call reaches the environment, the policy gate checks it
against ledger state. Thus, relevant state is exposed as a stable structured
representation rather than scattered across the transcript, and policy
constraints are re-checked at the action boundary.

\subsection{Ledger State and Updates}
\label{sec:method-ledger-state}
\label{sec:method-ledger-updates}

Task state is the snapshot of task-relevant facts, conditions, identifiers, and
data observed through interaction with the environment. The ledger stores the
portion observed through tools in a domain schema. It is not long-term memory, an
LLM summary, a per-task checklist, or a claim to recover unobserved world state.
Formally, the ledger is a typed dictionary $L: \mathcal{P}\rightarrow\mathcal{V}$,
where $\mathcal{P}$ is the set of canonical schema paths and $\mathcal{V}$ is the
set of tool-returned values. Paths are stable addresses for observed records,
such as \texttt{user}, \texttt{orders.*}, \texttt{products.*},
\texttt{reservations.*}, or keyed flight-search results. Nested values remain
inside stored records, and the model sees a deterministic rendering of the
ledger. Each domain supplies a fixed tool path map that routes whole successful
returns to canonical ledger locations. The map follows the tool interface and
policy-relevant entities; it is not generated by LLMs.

The ledger updates only from successful read-tool returns. For each return,
LedgerAgent links it to the earlier tool call, recovers the tool name and
arguments, parses the returned JSON, and stores the returned record at the path
determined by the map. Failed tools and write-tool returns do not update
states. After a successful write, the agent must
issue a read call to observe the new state. This observe-not-assume rule keeps
the ledger grounded in the external system. For example, in a retail exchange,
observed order and product records stay addressable under stable paths, so later
tool arguments and policy checks can use environment-returned identifiers rather
than searching earlier JSON in the transcript.

\subsection{Ledger-Grounded Generation}
\label{sec:method-generation}

Before each model call, LedgerAgent adds the full ledger block to the prompt. The
block is generated deterministically from $L$ and lists every record that has
actually been observed through read tools. Each entry is shown under a canonical
path, such as \texttt{orders.1234} or \texttt{products.5678}, together with the
stored returned value. The dialogue history, policy text, and
normal tool schemas are still provided; the ledger block is an additional state
view, not a replacement for them.

The purpose of the block is to make the current observed state easy for the model
to find. For example, after the agent reads an order and a product, the block
shows that \texttt{orders.1234} is delivered, belongs to the current user, and
contains item \texttt{sku\_a}, while \texttt{products.5678} lists the observed
replacement variants. If the user later says ``exchange that item,'' the model
can use these stable paths and identifiers instead of searching through earlier
JSON tool returns.

\subsection{Policy Gate}
\label{sec:method-policy-gate}

The policy gate runs immediately before any environment-changing call is executed.
It evaluates the proposed call against executable predicates $\Pi$ over the
current ledger $L$ and returns one of three outcomes:
\begin{itemize}[nosep]
    \item \textsc{allow}: execute the call unchanged.
    \item \textsc{revise}: remove the call and give the model the violated
    predicate.
    \item \textsc{block}: block the call and refuse the requested action.
\end{itemize}

For messages with multiple tool calls, the gate checks each environment-changing
call independently. Allowed calls remain in the assistant message; rejected calls
are removed. A \textsc{revise} verdict adds feedback to the next model turn, while
\textsc{block} ends the attempted environment-changing action.

The gate is only a verifier. It does not choose tools, repair arguments, fetch
missing records, or plan a new trajectory. The model still handles task progress;
the gate only checks whether a proposed action is consistent with the observed
ledger state and encoded policy.

Predicates are specified once per domain as code over ledger fields; we do not
compile arbitrary natural-language policy. In the reported experiments, the
policy layer contains 28 deterministic gate predicates in total: 10 for airline,
12 for retail, 6 for telecom, and none for telehealth. They encode recurring
checks such as ownership, entity-state preconditions, argument grounding, refund
or payment consistency, and loop prevention. Retail predicates check, for
example, that an order belongs to the authenticated user, that a return targets a
delivered order, that a refund uses an observed payment method, and that an
exchange item and replacement variant appear in observed records. Airline
predicates check that a selected flight came from a prior search result before it
is used in a reservation update.

Predicates use only records present in the ledger. If a rule requires explicit
evidence and that evidence is missing or inconsistent, the gate returns
\textsc{revise}. Otherwise, absence is not treated as a violation. Read calls are
allowed to repeat because re-reading is how the agent observes updated external
state.
Appendix~\ref{sec:appendix-running-example} gives a concrete trace of this
\textsc{revise} behavior.

\subsection{Agent Loop}
\label{sec:method-loop}

\begin{algorithm}[t]
\caption{LedgerAgent Loop}
\label{alg:ledgeragent}
\small
\begin{algorithmic}[1]
\Require message $m$, history $H$, ledger $L$, tools $T$, policy $P$, predicates $\Pi$
\State Append $m$ to $H$
\If{$m$ is a tool-return message}
    \State $L \gets \mathrm{Absorb}(L, m)$
    \Comment{successful known reads update typed state}
\EndIf
\State $C \gets \mathrm{Render}(L)$
\State $a \gets \mathrm{Generate}(H, P, C, T)$
\If{$a$ proposes environment-changing call(s)}
    \State $(a', g) \gets \mathrm{GateFilter}(a, L, \Pi)$
    \If{$g=\textsc{allow}$}
        \State \Return $a'$ \Comment{call(s) preserved unchanged}
    \ElsIf{$g=\textsc{revise}$}
        \State \Return $a'$ \Comment{rejected call(s) removed; feedback added}
    \ElsIf{$g=\textsc{block}$}
        \State \Return refusal for the requested environment-changing action
    \EndIf
\EndIf
\State \Return $a$
\end{algorithmic}
\end{algorithm}

LedgerAgent keeps one base-model generation per turn.
Algorithm~\ref{alg:ledgeragent} highlights the cost invariant in the default
configuration: ledger updates, ledger rendering, and policy checks wrap the base
model call, but they do not introduce an additional LLM call. Ledger updates
are deterministic operations over tool returns, rendering is deterministic string
formatting, and the policy gate is an executable predicate check over the typed
ledger. The model weights, tool schemas, and decoding procedure remain unchanged.

\begin{table}[!h]
\centering
\small
\begin{tabular}{llcl}
\toprule
\textbf{Domain} & \textbf{Benchmark} & \textbf{Tasks} & \textbf{Control} \\
\midrule
Airline & $\tau^{2}$-bench & 50 & single \\
Retail & $\tau^{2}$-bench & 114 & single \\
Telecom & $\tau^{2}$-bench & 114 & dual \\
Telehealth & $\tau$-Trait & 20 & single \\
\bottomrule
\end{tabular}
\caption{Benchmark domains. In single-control domains, only the agent modifies
the task database. In the dual-control setting, the user simulator can also
change shared state.}
\label{tab:experiment-domains}
\end{table}

To instantiate LedgerAgent in a new domain, a developer specifies two reusable
components: tool path maps for storing returned records, and executable
predicates for environment-changing tools. Both are domain-level, not task-level:
they follow the API records and policy constraints such as ownership,
entity state, grounded arguments, and payment consistency. The
method therefore applies to structured tool-use domains where compliance can be
checked against observed records.

\begin{table*}[t]
\centering
\scriptsize
\setlength{\tabcolsep}{3pt}
\resizebox{\textwidth}{!}{
\begin{tabular}{lcccccccccccc}
\toprule
\multirow{2}{*}{\textbf{Model}} 
& \multicolumn{3}{c}{$\tau$-Airline} 
& \multicolumn{3}{c}{$\tau$-Retail} 
& \multicolumn{3}{c}{$\tau$-Telecom} 
& \multicolumn{3}{c}{$\tau$-Telehealth} \\
\cmidrule(lr){2-4}
\cmidrule(lr){5-7}
\cmidrule(lr){8-10}
\cmidrule(lr){11-13}
& Avg & Pass\textasciicircum{}1 & Pass\textasciicircum{}4
& Avg & Pass\textasciicircum{}1 & Pass\textasciicircum{}4
& Avg & Pass\textasciicircum{}1 & Pass\textasciicircum{}4
& Avg & Pass\textasciicircum{}1 & Pass\textasciicircum{}4 \\
\midrule

Kimi-K2.5 (FC)
& 54.4\% & 69.0\% & 44.0\%
& 38.3\% & 57.5\% & 24.6\%
& 80.9\% & 90.8\% & 71.9\%
& 11.3\% & 15.0\% & 10.0\% \\

Kimi-K2.5 (Ledger)
& 62.3\% & 74.0\% & 52.0\%
& 53.9\% & 70.6\% & 41.2\%
& 69.9\% & 76.5\% & 64.0\%
& 18.8\% & 25.0\% & 15.8\% \\

\addlinespace[2pt]
\hdashline
\addlinespace[2pt]

GLM-5 (FC)
& 51.3\% & 66.5\% & 40.0\%
& 40.9\% & 61.0\% & 26.3\%
& 63.7\% & 80.3\% & 50.9\%
& 16.9\% & 20.0\% & 15.8\% \\

GLM-5 (Ledger)
& 64.6\% & 76.0\% & 56.0\%
& 48.5\% & 67.1\% & 35.1\%
& 68.7\% & 75.9\% & 62.3\%
& 17.6\% & 27.5\% & 10.0\% \\

\addlinespace[2pt]
\hdashline
\addlinespace[2pt]

MiniMax M2.5 (FC)
& 46.2\% & 61.5\% & 36.0\%
& 16.7\% & 33.6\% & 7.0\%
& 66.1\% & 81.8\% & 53.5\%
& 10.7\% & 18.8\% & 5.0\% \\

MiniMax M2.5 (Ledger)
& 49.9\% & 63.0\% & 40.0\%
& 36.6\% & 58.1\% & 21.1\%
& 66.3\% & 74.8\% & 58.8\%
& 20.7\% & 28.8\% & 15.0\% \\

\bottomrule
\end{tabular}
}
\caption{Main results comparing standard function calling (FC) with Ledger
across the evaluated $\tau^{2}$-Bench and $\tau$-Trait domains. For each
backbone and domain, Avg reports the average of pass\textasciicircum{}1 and
pass\textasciicircum{}4; higher values indicate better task success and
run-to-run consistency.}
\label{tab:main_results}
\end{table*}

\section{Experiments}
\label{sec:experiments}

We compare \textsc{LedgerAgent} against a standard prompt-based tool-calling
baseline on customer-service tasks 
from $\tau^{2}$-bench~\cite{barres2025tau2} and
$\tau$-Trait~\citep{he2025impatient}. Each task provides a user
goal, an initial database state, a domain policy, and tools; the agent must
complete the task through dialogue and tool use.

\subsection{Benchmark Domains}
\label{sec:experiments-domains}

Table~\ref{tab:experiment-domains} lists the four domains. Airline and retail
focus on grounding writes in retrieved records. Telecom tests dual-control
behavior, where user-side actions can also change state. Telehealth comes from
$\tau$-Trait and keeps the same structured tool-use format in a single-control
setting.

\subsection{Agent Conditions}
\label{sec:experiments-agents}

For each backbone, we compare the baseline agent with \textsc{LedgerAgent}. Both
conditions use the same policy, tools, conversation history, decoding settings,
and number of model calls. The baseline recovers task state from the transcript.
\textsc{LedgerAgent} additionally renders the observed ledger before generation
and, except in telehealth, checks proposed environment-changing calls with the
policy gate. Thus, the comparison isolates the ledger representation and action
boundary check rather than extra calls, tools, or training.

\subsection{Agent and User Models}
\label{sec:experiments-models}

We evaluate six agent models: GPT-5.2, GPT-4.1~\cite{achiam2023gpt}, Kimi K2.5~\cite{team2026kimi}, GLM-5~\cite{zeng2026glm},
MiniMax-M2.5~\cite{MiniMax2025MiniMax01SF}, and Qwen3-30B~\cite{Yang2025Qwen3TR}. For each model, we compare LedgerAgent against the
corresponding standard-agent baseline built from the same underlying model. 
Unless otherwise stated, all agent runs use temperature 0.0 and all
comparisons use the same fixed user simulator; across all experiments in
the paper, the user simulator is GPT-5-mini.

\subsection{Evaluation Protocol}
\label{sec:experiments-protocol}

We run four independent trials per task for each domain--model--agent cell. A
task receives pass\textasciicircum{}k if all $k$ trials pass. We report
pass\textasciicircum{}1 as the main success metric and pass\textasciicircum{}4 as
a consistency measure. Rewards are computed by the benchmark evaluator from the
task-specific database, action, communication, and natural-language checks.

\section{Results}
\label{sec:results}


\begin{table}[t]
\centering
\footnotesize
\setlength{\tabcolsep}{3pt}
\begin{tabular}{lccc}
\toprule
\textbf{Method} & \textbf{Pass\textasciicircum{}1 $\uparrow$} & \textbf{Pass\textasciicircum{}4 $\uparrow$} & \textbf{Token Overhead $\downarrow$} \\
\midrule
IRMA & 23.4\% & 9.6\% & 53.1\% \\
Ledger (ours) & 27.2\% & 17.1\% & 0.0\% \\
\bottomrule
\end{tabular}
\caption{Comparison of test-time agentic methods in terms of task success, repeated-run reliability, and token overhead.}
\label{tab:agentic-overhead}
\end{table}

\paragraph{Cross-Model Generalization}
Table~\ref{tab:main_results} reports results for three non-GPT backbone models.
Across these backbones, Ledger improves average performance over the standard
Function Calling (FC) baseline. With Kimi-K2.5 as the backbone, Ledger improves
over FC by 3.4 points in average pass\textasciicircum{}1 and 5.6 points in
average pass\textasciicircum{}4. The same trend holds for GLM-5 and MiniMax
M2.5: Ledger achieves gains of 4.7 and 7.3 points in average
pass\textasciicircum{}1, respectively, and 7.6 and 8.3 points in average
pass\textasciicircum{}4, respectively. These results show that Ledger is not
tied to a particular backbone model; instead, it provides consistent average
improvements across model families and task domains.

\begin{figure}[]
    \centering
    \includegraphics[width=\linewidth]{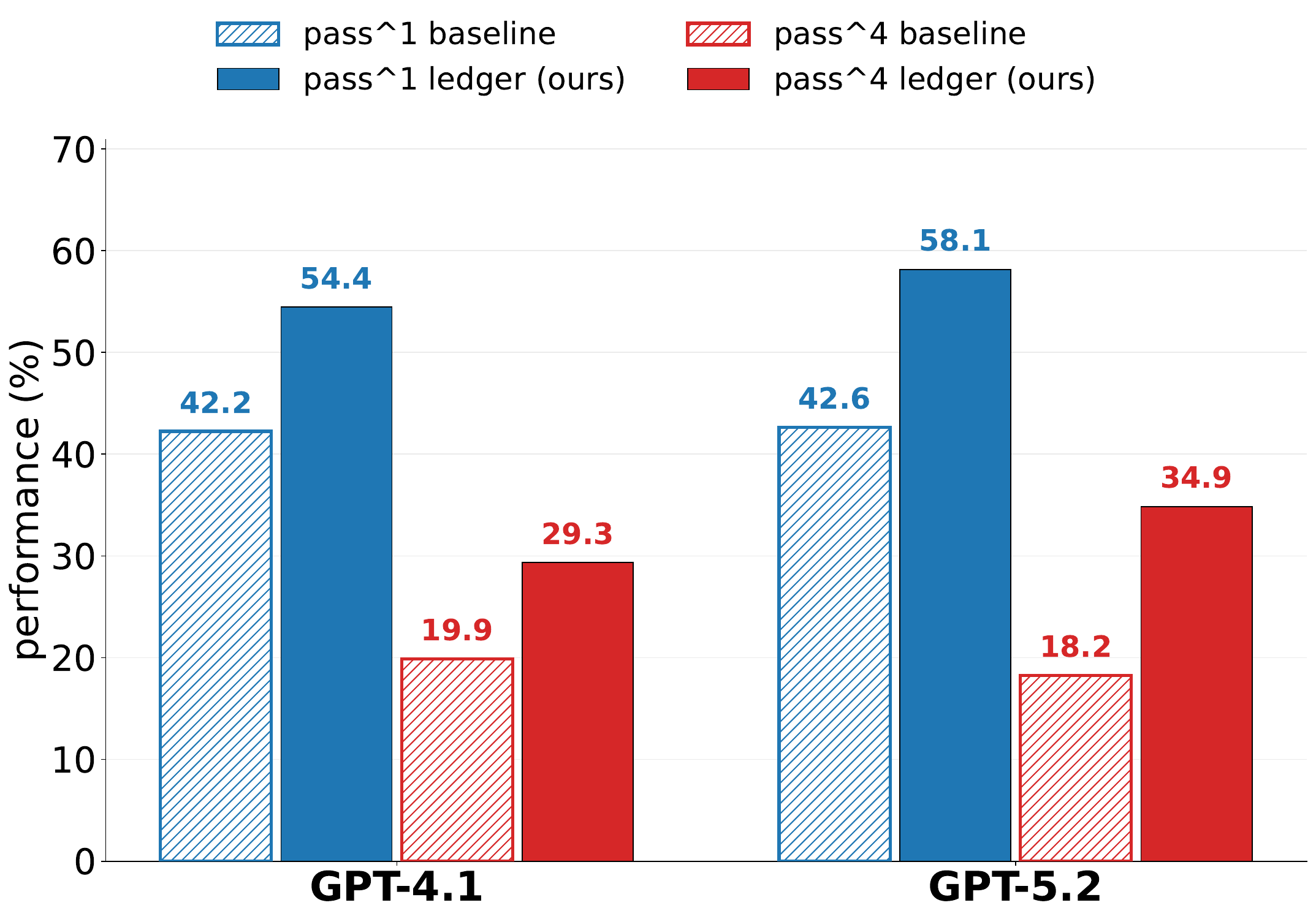}
    \caption{Pass\textasciicircum{}k results for GPT backbones. Higher
    pass\textasciicircum{}k indicates that the agent solves the same task more
    consistently across independent trials.}
    \label{fig:passk-bars-gpt}
\end{figure}

We further evaluate Ledger using GPT-based backbones. Because these models are
more costly, we restrict this comparison to the retail and airline environments.
As shown in Figure~\ref{fig:passk-bars-gpt}, Ledger outperforms the FC baseline
by 12.2 and 15.5 points in average pass\textasciicircum{}1 when using GPT-4.1
and GPT-5.2, respectively, as the backbone of the tool-calling agent. We observe
comparable improvements in pass\textasciicircum{}4. These results provide
additional evidence that Ledger improves both the accuracy and consistency of
tool-calling agents on complex tasks in dynamic environments.

\begin{figure*}[t]
    \centering
    \includegraphics[width=\linewidth]{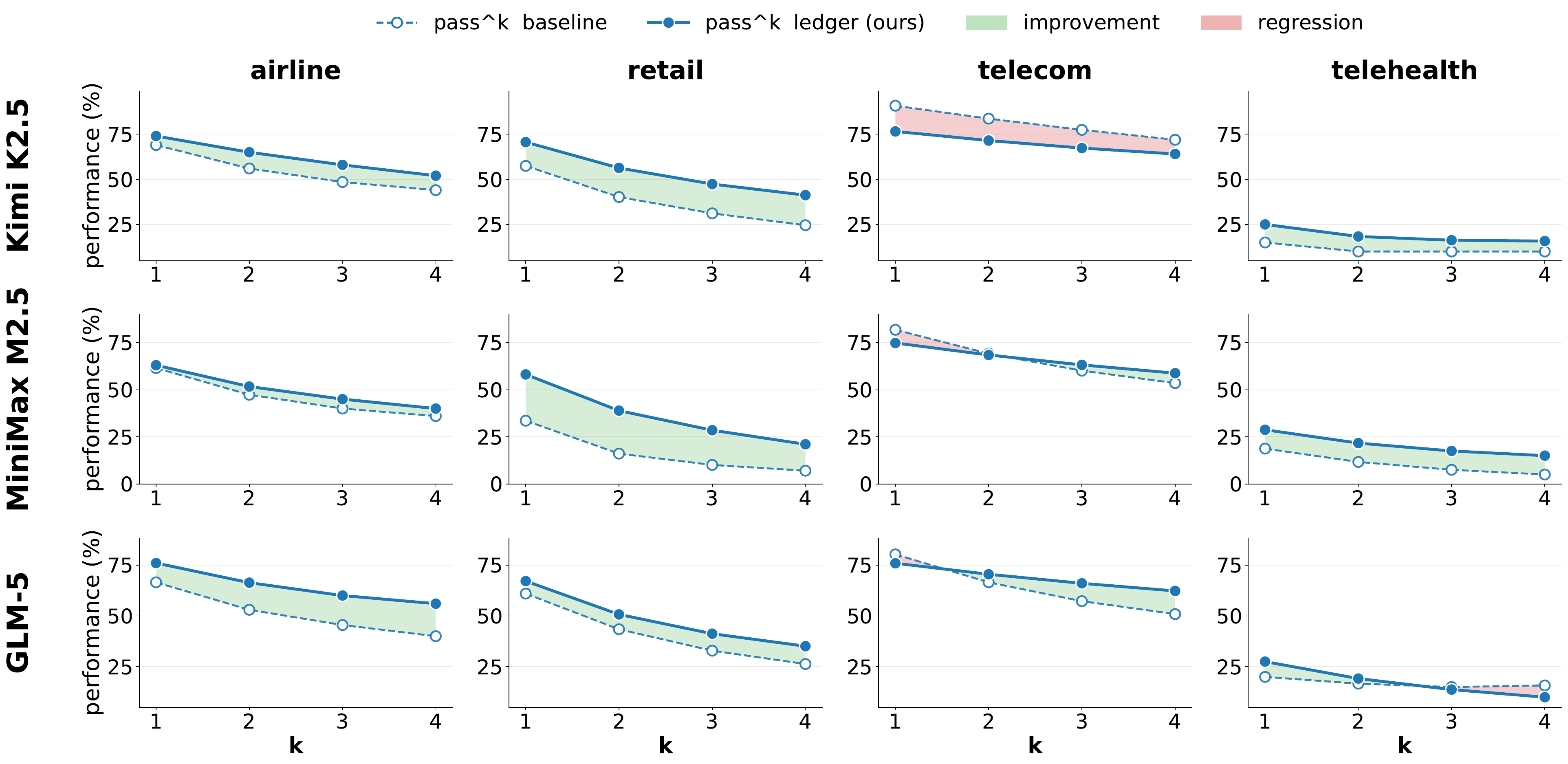}
    \caption{Performance on tasks that require at least one environment-changing
    tool call, defined as a tool call that modifies external system state, such
    as updating an order, issuing a refund, changing a reservation, or modifying
    an account. \textsc{LedgerAgent} improves both  pass\textasciicircum{}1 and
    pass\textasciicircum{}4 on this write-action subset, indicating better reliability
    when task success depends on modifying external state.}
    \label{fig:write_main_res}
\end{figure*}


\paragraph{Comparison with Agentic Context-Engineering Methods}
Recent agentic methods~\cite{mishra2025irma, saeidi2026fama} have shown strong
performance gains in tool-calling environments. To assess Ledger against this
class of approaches, we compare it with IRMA, a recent context-engineering
method. As reported in Table~\ref{tab:agentic-overhead}, Ledger outperforms IRMA
by 3.7 points in pass\textasciicircum{}1 and 7.4 points in
pass\textasciicircum{}4. Importantly, Ledger achieves these gains without
introducing additional token overhead, whereas IRMA incurs more than 50\% token
overhead due to its use of three helper agents. This result shows that Ledger
not only improves over conventional baselines such as FC, but also surpasses a
recent agentic context-engineering method while being substantially more
token-efficient.

\begin{figure}[h]
    \centering
    \includegraphics[width=\linewidth]{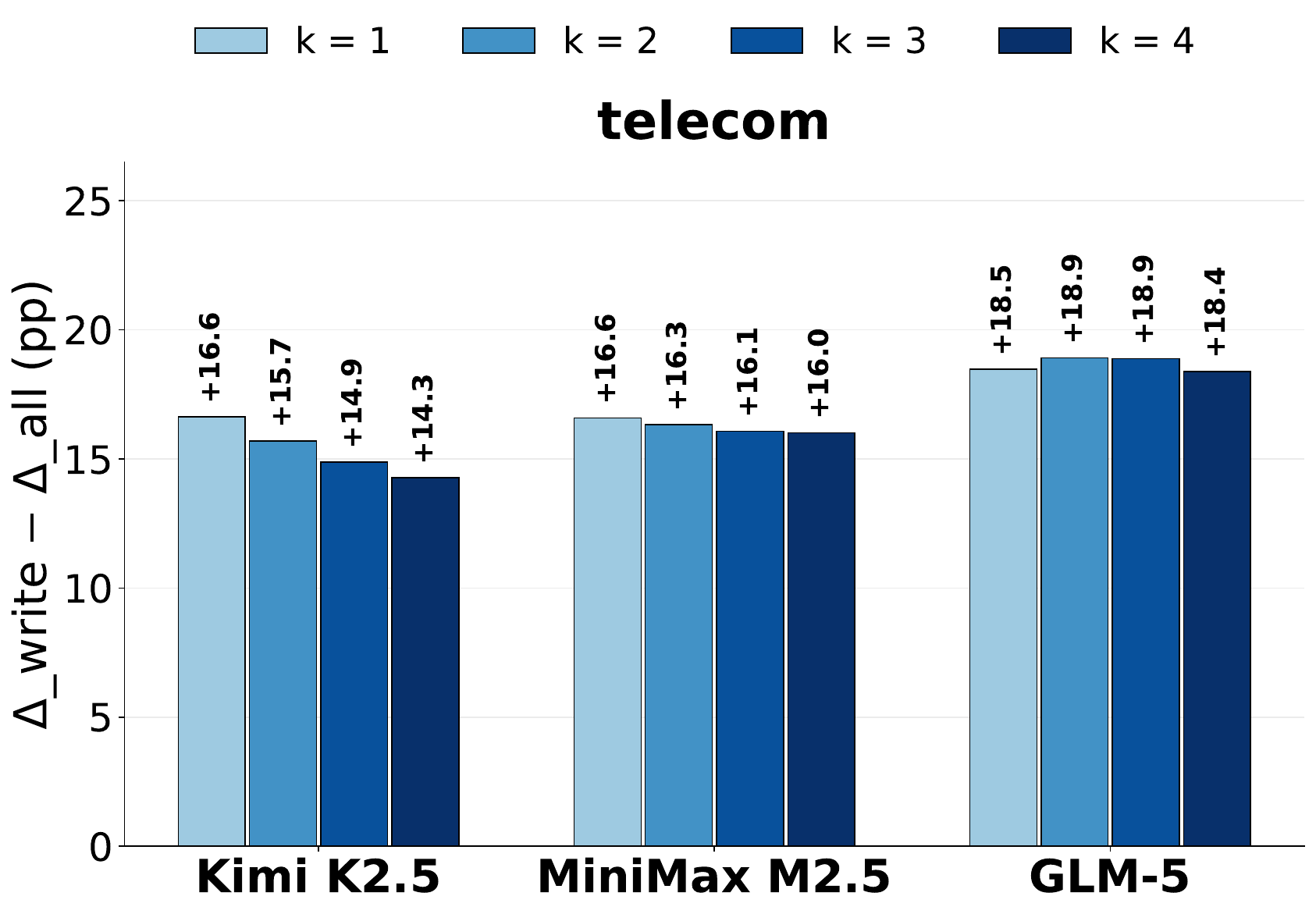}
    \caption{Telecom write-action results. The dual-control setting is especially
    sensitive to state drift because both the agent and user simulator can affect
    the shared database; Ledger improves action-level reliability by grounding
    proposed writes in the observed ledger state.}
    \label{fig:write_telecom_res}
\end{figure}

\paragraph{Performance on Environment-Changing Tasks}
A common failure mode in tool-calling agents occurs when an agent modifies
environment state and the change cannot be reverted. We refer to such operations
as environment-changing tool calls. To study this setting, we conduct a
comprehensive analysis of the evaluation tasks and identify tasks that require
at least one write action. We find that this category includes 26 out of 50
tasks in Airline, 104 out of 114 tasks in Retail, 94 out of 114 tasks in
Telecom, and 19 out of 20 tasks in Telehealth. We then evaluate both the
baseline methods and Ledger on this subset of tasks.

As shown in Figure~\ref{fig:write_main_res}, Ledger consistently outperforms the
baselines on tasks involving environment-changing tool calls. The improvement is
especially notable in the Telecom domain, where Ledger substantially increases
action-level performance compared with the baseline methods, as shown in
Figure~\ref{fig:write_telecom_res}. These results support our hypothesis that
Ledger improves the ability of tool-calling models to solve complex tasks that
require reliable state tracking and careful execution of write actions.

\begin{figure*}[t]
    \centering
    \includegraphics[width=\linewidth]{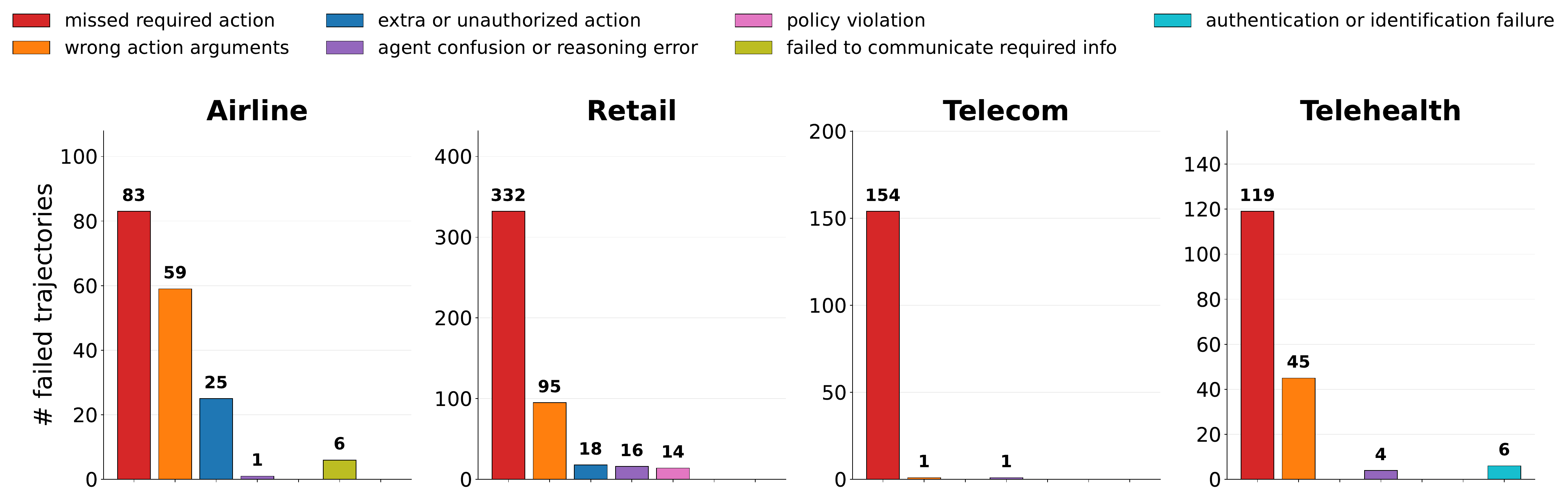}
    \caption{Failure categories for Ledger trajectories across domains and
    backbone models. Missed required actions dominate, with domain-specific
    wrong-argument and unauthorized-action errors.}
    \label{fig:error_analysis}
\end{figure*}

\section{Error Analysis}
\label{sec:error_analysis}

We analyze failed trajectories produced by three backbone models, Kimi K2.5, MiniMax M2.5, and GLM-5, under the Ledger setting across four domains (airline, retail, telecom and telehealth). As shown in Figure~\ref{fig:error_analysis}, missed required actions and wrong action arguments account for 90.7\% of all failures, with missed required actions representing 70.3\% and wrong action arguments representing 20.4\%. This distribution shows that Ledger agents fail much more often by omitting an expected tool call than by invoking the correct tool with incorrect arguments.

In many missed-action cases, the agent completes the initial lookup steps but terminates or transfers to human support when an edge case arises, such as a payment-method limitation, instead of continuing with the remaining policy-compliant write actions. The remaining 9.3\% of failures consists of extra or unauthorized actions, reasoning errors or tool-call loops, policy violations, communication failures, and authentication or identification failures. We assign these categories using a deterministic rule-based classifier based on evaluator reward signals, per-action match flags, and transcript evidence, including whether the expected tool was invoked and whether the agent transferred to human support.

The domain-level analysis shows that Ledger failures have different characteristics across environments. Retail failures are primarily missed actions, accounting for 69.9\% of failures, followed by wrong arguments at 20.0\%. These failures often occur when the agent agrees to a multi-item modification but transfers to human support after encountering a payment or eligibility constraint, rather than following a valid modify-payment or partial-fulfillment path. Telecom failures are almost entirely dominated by missed actions, which account for 98.7\% of failures. These cases typically involve the agent failing to invoke a required permission-grant tool or a required transfer step.

Airline has the most diverse failure profile, with missed actions accounting for 47.7\% of failures and wrong arguments for 33.9\%. It also contributes the largest number of extra or unauthorized actions, often when the agent concedes after user pushback and performs a policy-disallowed flight change or cabin downgrade. Telehealth has the highest wrong-argument rate, at 25.9\%, reflecting its more complex tool schemas, including fields such as \texttt{provider\_id}, \texttt{appointment\_type}, \texttt{bill\_insurance}, and \texttt{payment\_notes}. Telehealth is also the only domain with authentication or identification failures, where the agent transfers to human support without first attempting the available patient lookup. Overall, these results suggest that Ledger's remaining errors are domain-specific. Retail and Telecom require better handling of premature transfers, Airline requires stronger safeguards before irreversible writes, and Telehealth requires more robust schema-aware argument extraction.

\section{Conclusion}
\label{sec:conclusion}

We introduced \textsc{LedgerAgent}, an inference-time method for
policy-adherent tool-calling agents that makes observed task state explicit
when actions are chosen. The core problem is state grounding: agents may
retrieve the right records but still act on stale, missing, or incorrectly
reconstructed information when state remains buried in the transcript.
\textsc{LedgerAgent} addresses this with two deterministic components: a
schema-anchored ledger that stores successful read-tool returns as typed state,
and a policy gate that checks proposed environment-changing calls against that
state before execution.
Across customer-service domains and backbone models, \textsc{LedgerAgent}
improves policy-adherent tool use without changing model weights, with the
largest gains on consistency-oriented pass$^k$ metrics and tasks requiring
environment-changing actions. Error analysis shows that remaining failures are
mostly missed actions and domain-specific argument errors, suggesting that
explicit state and write-time verification reduce one important source of
unreliability but do not replace planning or schema-aware argument construction.
Overall, \textsc{LedgerAgent} supports a simple design principle: state that
determines whether an action is valid should be represented and checked
explicitly, rather than left only in an expanding prompt history.

\section{Limitations}
\label{sec:appendix-limitations}
\textsc{LedgerAgent} is designed for structured tool-use domains. It assumes
that tool returns expose stable fields that can be mapped into a domain schema,
which matches the customer-service settings studied here, such as users, orders,
reservations, accounts, and other record-like entities. The approach is less
direct for tasks where the relevant state is primarily unstructured, visual,
latent, or unavailable through read tools. The ledger also contains only observed
state. It cannot certify facts that the agent has not retrieved, and after an
environment-changing call the ledger reflects the new external state only after
that state is observed again through a read call. In such cases, the policy gate
can request additional evidence or abstain, but final success still depends on
the agent gathering the necessary observations.

The current implementation uses domain-level specifications. A developer defines
the read-tool path map and encodes recurring policy clauses as executable
predicates. These specifications are reusable across tasks in a domain and do
not require additional model training, but they are not automatic policy
induction. Consequently, \textsc{LedgerAgent} improves enforcement for covered,
observable constraints rather than providing a complete proof of policy
compliance for every possible interaction. Missing schema fields, ambiguous
policy language, or omitted predicates can still leave some errors to the
underlying model and benchmark evaluator.

Our empirical evaluation is also scoped to the benchmarks considered in this
work. We evaluate four customer-service domains with structured APIs and a fixed
user simulator, covering both single-control and dual-control settings but not
the full range of live users, adversarial behavior, changing policies, or
production traffic. We run four independent trials per task, which supports the
pass\textasciicircum{}4 consistency analysis but does not characterize very long
dialogues or rare failure modes. In addition, our comparison with agentic
context-engineering methods uses one representative method rather than an
exhaustive set of possible multi-agent or memory-based systems.

Finally, the default configuration keeps the same number of LLM calls as the
standard agent, but it is not cost-free. Rendering the ledger adds prompt
content, and maintaining schemas and predicates adds implementation and testing
overhead. These costs are most compelling when tool returns are structured and
environment-changing actions are governed by clear, recurring policy
constraints; they may be less justified for simple tasks where transcript-only
state tracking is already reliable.

\section*{Ethical Considerations}
\label{sec:appendix-ethics}
The authors state that this work is in accordance with the ACL Code of Ethics
and does not raise ethical issues. AI assistants, specifically Grammarly and
ChatGPT, were utilized to correct grammatical errors and restructure sentences.

\bibliography{anthology,custom}
\bibliographystyle{acl_natbib}

\appendix

\section{Licenses}
\label{sec:appendix-licenses}
This work uses models and benchmarks from various sources. Table~\ref{tab:licenses}
lists the licenses for the models and benchmarks used in this study.

\begin{table}[h]
\centering
\small
\begin{tabular}{ll}
\toprule
\textbf{Model/Benchmark} & \textbf{License Type} \\
\midrule
GPT-5.2 & Proprietary (OpenAI) \\
GPT-4.1 & Proprietary (OpenAI) \\
Kimi K2.5 & Modified MIT \\
GLM-5 & MIT \\
MiniMax M2.5 & Modified MIT \\
Qwen3-30B-Instruct & Apache 2.0 \\
$\tau^{2}$-Bench & MIT \\
$\tau$-Trait & MIT \\
\bottomrule
\end{tabular}
\caption{Licenses for models and benchmarks used in this study.}
\label{tab:licenses}
\end{table}
\section{Running Example 1: Gating a Policy-Violating Write}
\label{sec:appendix-running-example-block}

This trace shows the gate \emph{blocking} a non-compliant write rather
than correcting a recoverable one. It is taken from a single passing LedgerAgent
trajectory on $\tau^{2}$-Bench airline task 28 (reward $1.0$); all identifiers,
reservation fields, the gate verdict, and the verbatim rejection message come
from the evaluated environment. For this task the gold trajectory contains
\emph{no} write: the policy-correct behavior is to decline the cancellation, so
the gate's block is precisely what earns the reward.

\paragraph{Task setup.}
The user is Amelia Rossi, \texttt{amelia\_rossi\_1297}. She asks to cancel
reservation \texttt{SI5UKW} and receive a refund. The reservation is a
\texttt{basic\_economy} one-way trip MIA--PHX (via LAS) booked on 2024-05-11,
with no travel insurance. Under airline policy a non-business reservation may be
cancelled only within 24 hours of booking, or with travel insurance, or if the
airline cancelled a flight; none of these holds here. The ledger starts empty:
$L=\{\}$.

\paragraph{Conversation transcript.}
The full dialogue is reproduced below; tool returns are abbreviated to the
fields the gate consults, and a \textbf{Gate} line marks where the policy layer
intercepted a proposed write and returned its reason to the model.
{\footnotesize
\begin{flushleft}
\textbf{Agent:} Hi! How can I help you today?\\[3pt]
\textbf{User:} I need to cancel reservation \texttt{SI5UKW} and get a refund. I'm
Amelia Rossi (user id \texttt{amelia\_rossi\_1297}); refund to the original form
of payment. I do not accept ``no'' --- if a cash refund isn't possible I want a
full-value voucher, otherwise 50\%, otherwise at least 10\% with fees waived. If
you can't process this now, escalate me to a supervisor immediately.\\[3pt]
\textbf{Agent} \textit{(read):} \path|get_reservation_details(SI5UKW)| $\rightarrow$
\path|ledger.reservations.SI5UKW| \texttt{\{cabin: basic\_economy, MIA$\to$PHX,
created\_at: 2024-05-11, insurance: no\}}\\[3pt]
\textbf{Agent} \textit{(read):} \path|get_user_details(amelia_rossi_1297)|
$\rightarrow$ \path|ledger.user|\\[3pt]
\textbf{Gate} \textnormal{(\textsc{block})}: drops
\path|cancel_reservation(SI5UKW)| --- ``Per airline policy, reservation SI5UKW
(basic\_economy) cannot be cancelled outside the 24-hour booking window without
travel insurance and without an airline-cancelled flight.''\\[3pt]
\textbf{User:} Don't close this out. Either add trip insurance retroactively and
then cancel, or escalate me to a human supervisor now; and send me the exact
policy citation.\\[3pt]
\textbf{Agent} \textit{(read):} \path|get_reservation_details(SI5UKW)| (re-reads;
typed state unchanged)\\[3pt]
\textbf{Agent:} The booking is basic economy, outside the 24-hour window, and
uninsured; insurance cannot be added retroactively and the cancellation cannot be
processed. Escalating to a human supervisor as requested. \texttt{[\#\#\#STOP\#\#\#]}
\end{flushleft}}

Table~\ref{tab:running-example-block-trace} summarizes the key turns in this
blocking example.
\begin{table*}[t]
\centering
\footnotesize
\setlength{\tabcolsep}{3pt}
\begin{tabular}{p{0.06\textwidth}p{0.24\textwidth}p{0.34\textwidth}p{0.28\textwidth}}
\toprule
\textbf{Turn} & \textbf{Event} & \textbf{Ledger operation} & \textbf{Policy consequence} \\
\midrule
2 & Agent reads the reservation and the user record. & The path rules store \texttt{SI5UKW} under \texttt{ledger.reservations} and the customer object at \texttt{ledger.user}. The reservation shows cabin \texttt{basic\_economy}, booking date 2024-05-11, and insurance \texttt{no}. & The gate now has the typed fields needed to evaluate cancellation eligibility. \\
3 & Model proposes a cancellation. & No state change occurs; the call is intercepted before execution. & The gate returns \textsc{block}: the reservation is basic economy, outside the 24-hour window, uninsured, and has no airline-cancelled flight. The requested write is refused. \\
4 & Lacking any qualifying basis, the agent does not reissue the cancellation; it explains the policy. The user demands a supervisor override. & No write is recorded; \texttt{ledger.reservations.SI5UKW} is unchanged. & The agent holds the policy and ends the conversation. The reservation is untouched, matching the gold trajectory; reward $1.0$. \\
\bottomrule
\end{tabular}
\caption{Turn-level trace for task 28. The agent attempts a cancellation that
violates policy; the gate reads the typed reservation fields and blocks the
call. Because the policy-correct action is to refuse, leaving the database
unchanged earns the reward.}
\label{tab:running-example-block-trace}
\end{table*}

\paragraph{Read absorption.}
As before, each read lands at a canonical typed path through a domain-level path
rule. The agent calls \path|get_reservation_details| for \texttt{SI5UKW}, stored
at \path|ledger.reservations.SI5UKW|, and \path|get_user_details| for the
customer, stored at \path|ledger.user|. The reservation's typed state records a
\texttt{basic\_economy} one-way itinerary MIA--PHX with a single passenger:
\texttt{HAT062} (MIA--LAS, 2024-05-16) and \texttt{HAT284} (LAS--PHX,
2024-05-17); a creation timestamp of 2024-05-11; payment history of \$124 on
\texttt{credit\_card\_4579924}; and \texttt{insurance = no}. These are exactly
the fields the cancellation predicate consults.

\paragraph{Gate check and the block.}
The model then proposes the write the user asked for:
\begin{quote}
\small
\begin{tabular}{@{}l@{}}
\texttt{cancel\_reservation(}\\
\texttt{reservation\_id=SI5UKW)}
\end{tabular}
\end{quote}
Before the call reaches the environment, the gate evaluates the predicates
attached to \path|cancel_reservation| against the ledger. The ownership
predicate confirms \texttt{SI5UKW.user\_id} matches the observed user. The
decisive predicate is \texttt{cancel\_requires\_basis}, which reads typed fields
only: the cabin is \texttt{basic\_economy} (not business); \texttt{insurance} is
\texttt{no}; the booking timestamp 2024-05-11 is well outside the 24-hour window
relative to the evaluation reference date; and no flight in the reservation has
an airline-cancelled or delayed status in \texttt{ledger.flight\_status}. With
none of the four qualifying conditions met, the predicate blocks the call and
returns the verbatim reason:
\begin{quote}
\small
``Per airline policy, reservation SI5UKW (basic\_economy) cannot be cancelled
outside the 24-hour booking window without travel insurance and without an
airline-cancelled flight. Either confirm the user wants a transfer to human, or
add insurance first if eligible.''
\end{quote}
The offending \path|cancel_reservation| call is refused and the reason is
returned to the model. Because no qualifying basis exists, the agent does not
reissue the cancellation; it explains the policy to the user and, when the user
demands a supervisor override, holds the policy and ends the conversation. The
reservation is never cancelled, so the database is left in its correct state and
the trajectory is rewarded $1.0$.

This example highlights three points. First, the eligibility decision is made
over typed ledger fields --- cabin, booking timestamp, insurance flag, and
flight status --- not over the transcript, so the same predicate generalizes
across phrasings and across user pressure. Second, the gate enforces the policy
\emph{at the write boundary}: the non-compliant \path|cancel_reservation| is
intercepted before it can mutate the environment, rather than being caught after
the fact. Third, on a task whose policy-correct outcome is a refusal, this
interception is exactly what produces the correct, rewarded behavior, and it
does so even when the user explicitly presses the agent to override policy.

\section{Running Example 2: A Corrective Revision}
\label{sec:appendix-running-example}

This appendix gives a concrete execution trace for one real $\tau^{2}$-Bench
retail task, task 83. All user identifiers, order identifiers, item identifiers,
prices, the gate verdict, and the verbatim rejection message are taken from a
single passing LedgerAgent trajectory in the evaluated environment (reward
$1.0$). The example illustrates how the ledger is built from read tools, how the
gate checks a proposed write against typed state, and how a recoverable
\textsc{revise} verdict redirects a non-compliant argument so the task can
complete correctly.

\paragraph{Task setup.}
The user is Chen Silva, \texttt{chen\_silva\_7485}. A delivered order
\texttt{\#W9571698} contains a gold 128GB tablet (item \texttt{6065192424},
\$989.70). The user wants to return the tablet and have the \$989.70 refunded to
their Mastercard ending \texttt{2732} (\texttt{credit\_card\_1565124}). Under
retail policy, however, a return must be refunded to the order's original
payment method or to a gift card already in the user's profile. The order was
paid with gift card \texttt{7250692}, so the requested Mastercard refund is not
permitted. The ledger starts empty: $L=\{\}$.

\paragraph{Conversation transcript.}
The full dialogue is reproduced below; tool returns are abbreviated to the
fields the gate consults, and a \textbf{Gate} line marks where the policy layer
intercepted a proposed write and returned its reason to the model.
{\footnotesize
\begin{flushleft}
\textbf{Agent:} Hi! How can I help you today?\\[3pt]
\textbf{User:} I got two tablets in my delivery but only need one. I want to
return the more expensive one and have the refund go back to my credit card. I'm
Chen Silva, zip 46281; I don't remember my email or order number.\\[3pt]
\textbf{Agent} \textit{(reads):} \path|find_user_id_by_name_zip(Chen Silva, 46281)|
$\rightarrow$ \texttt{chen\_silva\_7485}; then \path|get_user_details| and
\path|get_order_details| on all four orders $\rightarrow$ \path|ledger.user|,
\path|ledger.orders.*| (order \texttt{\#W9571698}: \texttt{status: delivered},
gold 128GB tablet \texttt{6065192424} \$989.70, original payment
\texttt{gift\_card\_7250692}).\\[3pt]
\textbf{Agent:} The more expensive one is the gold 128GB tablet from order
\texttt{\#W9571698} (\$989.70). I'll refund \$989.70 to your Mastercard ending
2732. Do you confirm?\\[3pt]
\textbf{User:} Yes --- return the gold 128GB tablet from \texttt{\#W9571698} and
refund \$989.70 to my Mastercard ending 2732. \textit{(plus questions about email,
timing, return label, and fees)}\\[3pt]
\textbf{Gate} \textnormal{(\textsc{revise})}: drops
\path|return_delivered_order_items(..., payment_method_id=credit_card_1565124)|
--- ``Per retail policy, refunds must go to the original payment
(\texttt{['gift\_card\_7250692']}) or an existing gift card
(\texttt{['gift\_card\_7250692']}). You chose \texttt{'credit\_card\_1565124'}.''\\[3pt]
\textbf{User:} You can't refund to my Mastercard? That's ridiculous, but fine ---
refund the \$989.70 to the gift card on file (\texttt{gift\_card\_7250692}).\\[3pt]
\textbf{Agent} \textit{(write):}
\path|return_delivered_order_items(#W9571698, item_ids=[6065192424],|
\path|payment_method_id=gift_card_7250692)| $\rightarrow$ allowed; logged to
\path|ledger.history|\\[3pt]
\textbf{Agent:} Return processed: order \texttt{\#W9571698} is now ``return
requested'' and \$989.70 will be credited to gift card \texttt{7250692}.\\[3pt]
\textbf{User:} That works. Thanks for sorting it out. \texttt{[\#\#\#STOP\#\#\#]}
\end{flushleft}}

\begin{table*}[t]
\centering
\footnotesize
\setlength{\tabcolsep}{3pt}
\begin{tabular}{p{0.06\textwidth}p{0.24\textwidth}p{0.34\textwidth}p{0.28\textwidth}}
\toprule
\textbf{Turn} & \textbf{Event} & \textbf{Ledger operation} & \textbf{Policy consequence} \\
\midrule
2 & Agent authenticates the user and reads the order. & The path rules store the customer object at \texttt{ledger.user} and order \texttt{\#W9571698} under \texttt{ledger.orders}; the order shows status \texttt{delivered} and an original payment of gift card \texttt{7250692}. & The gate now has the order's payment history and the user's profile payment methods. \\
4 & Model proposes the return, refunding to the user's Mastercard \texttt{credit\_card\_1565124}. & No state change occurs; the call is intercepted before execution. & The gate returns \textsc{revise}: the chosen payment is neither the order's original payment nor a profile gift card. The call is dropped and the reason is returned. \\
5 & Model relays the policy; the user accepts a refund to the original gift card. The agent resubmits the return with \texttt{gift\_card\_7250692}. & The allowed write is recorded in \texttt{ledger.history}. & All predicates allow; the return executes and the trajectory is rewarded $1.0$. \\
\bottomrule
\end{tabular}
\caption{Turn-level trace for task 83. Read tools populate typed ledger paths;
the proposed refund is checked against the order's recorded payment history. The
refund predicate issues a recoverable \textsc{revise}; the agent redirects the
refund to the original payment method and the resubmission is allowed.}
\label{tab:running-example-trace}
\end{table*}

\paragraph{Read absorption.}
The path map contains one routing rule per read tool, defined at the domain
level and following the tool schema; the rules are not written for this task.
Here \path|find_user_id_by_name_zip| resolves the user id,
\path|get_user_details| is routed to \path|ledger.user|, and
\path|get_order_details| is routed to \path|ledger.orders.<id>|.

The customer object for Chen Silva is stored at \texttt{ledger.user}; its typed
fields include two payment methods, gift card \texttt{7250692} (balance \$59.0)
and Mastercard \texttt{credit\_card\_1565124} ending \texttt{2732}. The agent
then reads \texttt{\#W9571698}, stored at \path|ledger.orders.#W9571698|. The
typed state records status \texttt{delivered}; four delivered items including the
gold 128GB tablet \texttt{6065192424} at \$989.70; and a payment history showing
the order was paid with gift card \texttt{7250692}. The order's payment history
is exactly the field the refund predicate consults.

\paragraph{Gate check and the corrective revision.}
After the user confirms the return, the model proposes the write with the refund
directed to the user's Mastercard:
\begin{quote}
\small
\begin{tabular}{@{}l@{}}
\texttt{return\_delivered\_order\_items(}\\
\texttt{order\_id=\#W9571698,}\\
\texttt{item\_ids=[6065192424],}\\
\texttt{payment\_method\_id=}\\
\texttt{credit\_card\_1565124)}
\end{tabular}
\end{quote}
The gate evaluates the predicates attached to \path|return_delivered_order_items|
against the ledger. Ownership holds, the order is \texttt{delivered}, the item
belongs to the order, and the chosen payment is a valid profile method, so those
predicates allow. The decisive predicate is
\path|retail_refund_payment_in_order_history|, which reads the order's
recorded payment history and the user's profile gift cards and finds that the
chosen \texttt{credit\_card\_1565124} is neither. It returns \textsc{revise}
with the verbatim reason:
\begin{quote}
\small
``Per retail policy, refunds must go to the original payment
(\texttt{['gift\_card\_7250692']}) or an existing gift card
(\texttt{['gift\_card\_7250692']}). You chose \texttt{'credit\_card\_1565124'}.''
\end{quote}
Unlike a \textsc{stop} verdict, which terminates the episode, a \textsc{revise}
drops only the offending call and returns the reason to the model, which keeps
its turn. Note that the Mastercard \emph{is} a valid profile payment method, so
the generic profile check passes; it is the refund-specific provenance rule,
reading the order's payment history, that catches the violation.

\paragraph{Re-grounding and completion.}
The agent relays the constraint and, once the user agrees to a refund to the
original gift card, resubmits the identical return with
\path|payment_method_id=gift_card_7250692|. The refund predicate now finds
the destination in the order's payment history, every other predicate allows,
and the write executes; the event is appended to \texttt{ledger.history} and the
trajectory is rewarded $1.0$.

This example shows four properties of the implementation. First, construction is
automatic: each successful read lands at a canonical typed path through a
domain-level path rule, with no task-specific wiring. Second, predicates read
typed fields rather than transcript text; the revision at Turn~4 is a
membership test of the proposed refund destination against the order's recorded
payment history, not an LLM judgment over prose. Third, the gate enforces the
refund policy at the write boundary, redirecting a non-compliant refund the user
explicitly requested before it can mutate the environment. Fourth, the
\textsc{revise} verdict is corrective rather than terminal: it returns a typed
reason, the agent supplies a compliant argument, and the task completes.

\label{sec:appendix}

\end{document}